\documentclass{article}
\usepackage{spconf,amsmath,graphicx}
\usepackage{xcolor}
\title{COMPACT AND DE-BIASED NEGATIVE INSTANCE EMBEDDING FOR MULTI-INSTANCE LEARNING ON WHOLE-SLIDE IMAGE CLASSIFICATION}
\name{Joohyung Lee$^{\star}$, Heejeong Nam$^{\star}$, Kwanhyung Lee$^{\star}$, Sangchul Hahn$^{\star}$}
\address{$^{\star}$AITRICS, Republic of Korea} 
\begin{document}
%\ninept
%
\maketitle
%
% 과거 ICASSP REJECT 사건에서 교훈을 얻으면:
% 1) 너무 바이오스럽게 쓰지 말고 methodological novelty 최대한 있어보이게 쓰자 (결과 말고 methodological fancy). icassp은 바이오 학회라기보다 methodology 학회이다
% 2) (methodological 하게) 자신감있게 쓰자. 쭈구리 같이 썼다가 후회했었음.
% 3) marginal improvement 안된다
% For this signal processing conference the paper is of limited interest. Probably the paper would be better off by submitting it to a conference on applied machine learning with a focus on biomedical applications. This is also clear from the references that are a bit thin on signal processing society related papers.

% I believe the results of this paper are beneficial for medical applications, and practically, this is an interesting study. However, in terms of methodological novelty, it is not significant. Therefore, I keep my overall evaluation marginal accept.

% deepsvd 는 각 instance 기준으로 거리를 줄이는 center loss와 비슷함. 그런데 원 임. 이에 반해 우리는 feature dim-wise하게 stddev을 줄임

\begin{abstract}

Whole-slide image (WSI) classification is a challenging task because 1) patches from WSI lack annotation, and 2) WSI possesses unnecessary variability, e.g., stain protocol. Recently, Multiple-Instance Learning (MIL) has made significant progress, allowing for classification based on slide-level, rather than patch-level, annotations. However, existing MIL methods ignore that all patches from normal slides are normal. Using this free annotation, we introduce a semi-supervision signal to de-bias the inter-slide variability and to capture the common factors of variation within normal patches. Because our method is orthogonal to the MIL algorithm, we evaluate our method on top of the recently proposed MIL algorithms and also compare the performance with other semi-supervised approaches. We evaluate our method on two public WSI datasets including Camelyon-16 and TCGA lung cancer and demonstrate that our approach significantly improves the predictive performance of existing MIL algorithms and outperforms other semi-supervised algorithms. We release our code at https://github.com/AITRICS/pathology\_mil.

\end{abstract}
\section{Introduction}
\label{sec:intro}
Whole slide images (WSI) are digitized histology slides that preserve the bountiful information of original histology \cite{chen2023classifying,shao2021transmil}. With WSI, pathologists no longer use glass slides with a microscope but can apply various computing tools to process the digitized image \cite{li2023novel}. WSI has not only improved the workflow and diagnostic procedure of pathologists but also enabled disease analysis, e.g., cancer prognosis analysis, to benefit from deep learning technology. However, two main difficulties hinder the application of the conventional deep learning methods to whole slide image analysis.

% patch-level (or pixel-level) annotation is usually missing for WSI. 
First, WSIs are extremely large in size, often reaching up to $10^{10}$ pixels \cite{chen2023classifying}. Due to its exceptionally large size, existing methods usually employ patch-based processing - a WSI is split into small image patches which are then processed using the image encoder. However, a thorough examination of the whole slide image for a patch/pixel-level annotation is nearly infeasible due to the exceptionally large size of WSI, and therefore, a label for each image patch is usually lacking. Recently, multiple-instance learning (MIL) has been drawing attention in disease analysis of whole slide images since it exempts costly pixel-level annotation, which was required for traditional machine learning (ML) approaches for WSI analysis \cite{li2021dual}. Among various approaches within the MIL algorithm, attention-based MIL has shown promising results recently, which aggregates all instance embeddings into a single bag-level embedding using an attention module \cite{li2021dual, zhang2022dtfd, ilse2018attention}.

 % a label for each image patch is usually lacking. 
% First, WSIs are extremely large in size (typically reaching $10^{10}$ pixels) \cite{chen2023classifying}, and most regions are disease-negative even in disease-positive WSI \cite{li2021dual}. Due to its exceptionally large size, existing methods usually employ a patch-based method 

% pre-trained patch encoding using pre-trained weight and the second stage for proposing the MIL algorithm. 

% WSI often reaches up to $10^{10}$ pixels \cite{chen2023classifying}; consequently, a thorough examination of the whole slide image for an exhaustive patch/pixel-level annotation is nearly infeasible. Recently, multiple-instance learning (MIL) has been drawing attention in disease analysis of whole slide images since it exempts costly pixel-level annotation, which was required for traditional machine learning (ML) approaches for WSI analysis \cite{li2021dual}.

\begin{figure*}[t]
\centering
\includegraphics[width=\textwidth]{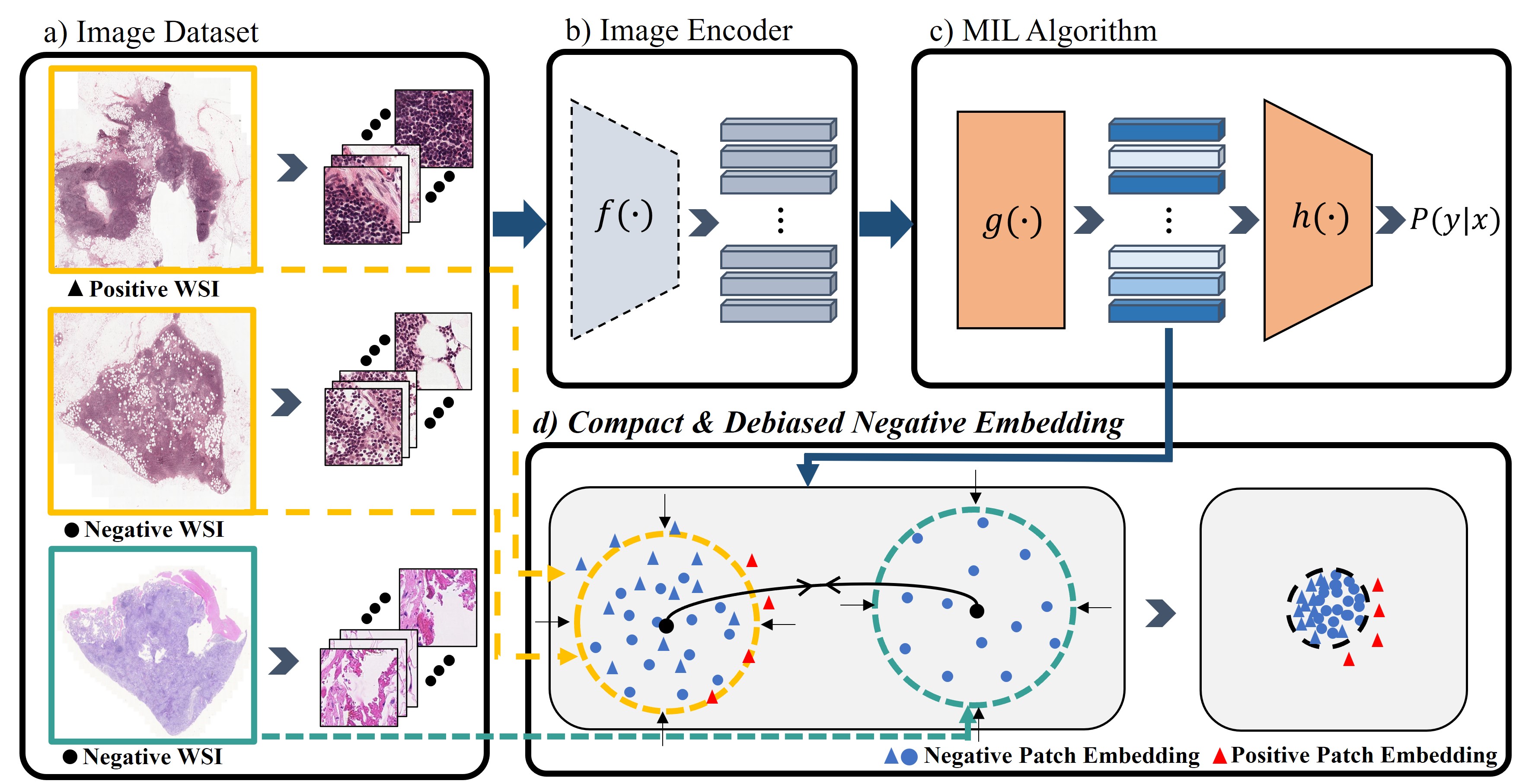}
\caption{Overview of our proposed \textbf{Compact and Debiased Negative Embedding} attached to the MIL algorithm. In this figure, the staining protocol is exemplified as the `bag-prior'; WSIs with darker stains are in yellow boxes whereas the green box indicates the WSI with lighter staining. Our method aims 1) to reduce the `bag-prior' by synchronizing the center of all patch embeddings from negative WSIs and 2) to impose a compact embedding for all patch embeddings from negative WSIs, as indicated by the black arrows. Note that triangles and dots in (d) are instance embeddings from different WSI indicated in (a).}
\label{fig1}
\end{figure*}

\begin{figure}
\centering
\includegraphics[width=0.48\textwidth]{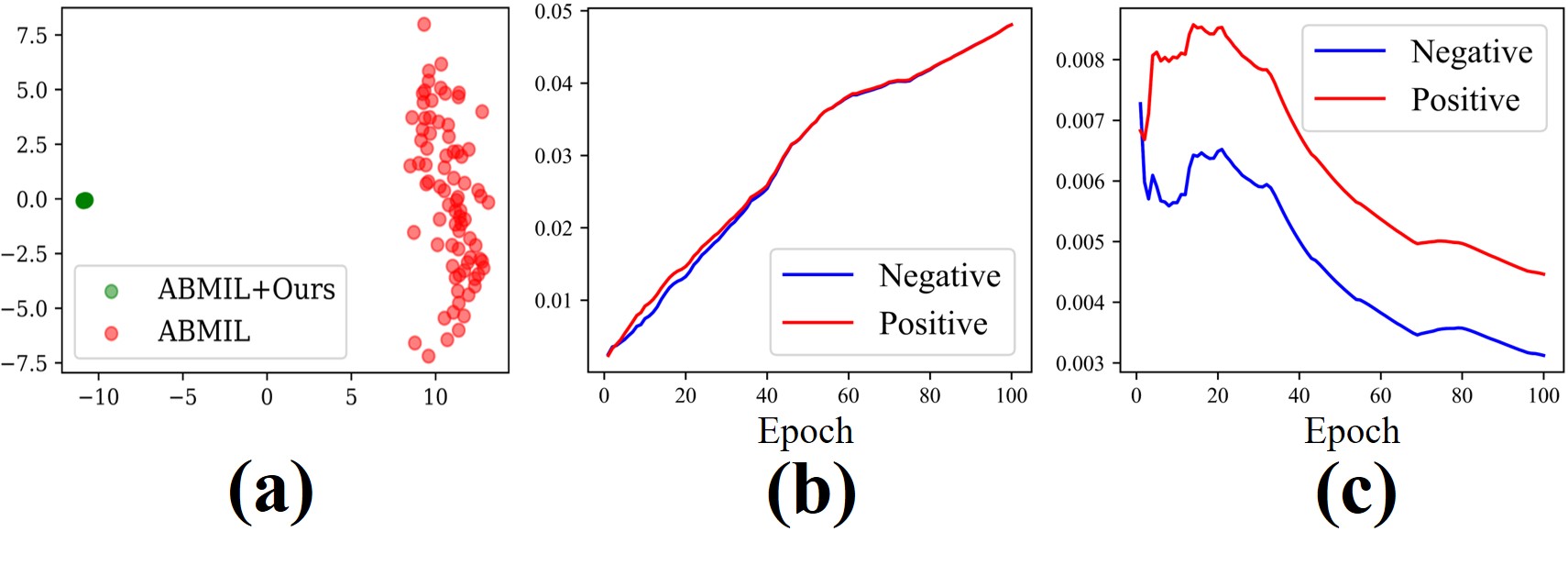}
\caption{(a) Instance embedding center per normal test WSIs (80 dots in both green and red); Mean standard deviation of instance embeddings per validation WSIs changes in ABMIL (Gated Attention) without ours (b); and with ours (c).}
\label{fig2}
\end{figure}

Second, WSIs possess wide variations, such as differences in tumor types, pen ink, and staining protocols \cite{wang2022scl, lin2023interventional, lin2022interventional, ianni2020tailored}. Though some variations can be informative, some variations are not and may even create a spurious correlation to the label. Lin \textit{et al.} \cite{lin2022interventional} has introduced a concept of `bag prior' and defined it as `an instance(patch)-shared information per bag (WSI) but irrelevant to the label', which therefore can cause the spurious correlation between instances and labels. In this study, we suggest improving the predictive performance of existing MIL algorithms by reducing the inter-slide bias by implementing a synchronized instance embedding center. The method and the result are depicted in Fig. \ref{fig1} and \ref{fig2}-(a), respectively.

% However, WSIs are extremely large in size, often reaching up to $10^{10}$ pixels \cite{chen2023classifying}. Given the overwhelming size of WSI, pixel-level annotation is not feasible, and the volume of data is too cumbersome for end-to-end learning. Recently, multiple-instance learning (MIL) has attracted increasing attention to learning WSI in a weakly-supervised manner. MIL does not require a pixel-level (instance) annotation but only demands slide-level (bag) annotation; MIL commonly segments WSI into smaller, manageable-size patches to enable machine learning processing.

The lack of local annotation for WSI is no longer problematic due to the widespread usage of the MIL algorithm. However, it has to be noted that WSI implicitly provides local annotation for negative slides; every tissue region from negative slides is negative. Therefore, even without explicit local annotation, we have access to normal-label (disease-negative) patches, which are disregarded in MIL. We aim to use this implicit information from WSI.

Such circumstance has been studied in one-class classification (OCC), where only a single class data is present during the training phase; data from another class, i.e., anomalous data, can appear during the test phase. Various approach has been introduced including the generative approach, synthesizing anomalies, and embedding-based approach \cite{liu2023simplenet}. Motivated by the embedding-based approach, we hypothesize that mapping all normal patches into a compact space can not only reduce the aforementioned inter-slide bias but also facilitate the model to capture the common factors of variation within the normal patches. Capturing common factors within normal patches can help distinguish normal, i.e. negative, and positive patches. We thus aim to demonstrate that creating compact normal patch embedding improves the predictive performance of existing MIL algorithms.

In this study, we aim to enhance the slide-level predictive accuracy of existing MIL methods using patches from the disease-negative slides. Specifically, we propose to minimize the standard deviation of all instance embeddings (before the attention aggregator) from negative WSI using a single learnable negative instance mean embedding vector. We hypothesize that compact negative representation will lead to more precise attention thus leading to the improved predictive performance of MIL algorithm.

We evaluate our method on two public WSI datasets, i.e., Camelyon-16 and TCGA lung cancer, to demonstrate that our approach improves the slide-level prediction of the existing MIL algorithms. We also compare the performance gain of our algorithm with other semi-supervised algorithms in WSI classification using MIL as well.

\section{Method}
\label{sec:method}

% Our aim is to facilitate the early stage of the MIL algorithm ($g(\cdot)$ in Fig. \ref(fig1)) to capture the common factors of variation within the normal patches, which can help assign a more precise attention score to each instance. Note that our method is orthogonal to the existing MIL algorithm and thus can be attached to any existing MIL algorithm to improve the predictive performance.

In this paper, we propose a semi-supervision signal that utilizes all patches from disease-negative slides to improve the predictive performance of the MIL model. Our aim is 1) to reduce the bias of each WSI, e.g., caused by varying staining protocol, and 2) to facilitate the mapping module in MIL algorithms, i.e., $g(\cdot)$ in Fig. \ref{fig1}, to capture the common factors of variation within the normal patches, which can help assign a more precise attention score to each instance in attention-based MIL. Note that our method is orthogonal to the existing MIL algorithms and thus can be attached to any existing MIL algorithms to improve the predictive performance.

% In this paper, we propose a semi-supervised signal using all patches from disease-negative slides to enhance MIL model accuracy. Our goals are 1) to mitigate WSI bias of each WSI, e.g., caused by varying staining protocol, and 2) to refine the MIL mapping module, i.e., $g(\cdot)$ in Fig. \ref{fig1}, for capturing common factors in normal patches, thereby improving attention scores. Note that our method is orthogonal to the existing MIL algorithms and thus can be attached to any existing MIL algorithms to improve the predictive performance.

\vspace{3mm}
\noindent\textbf{Definition 1} (WSI Dataset). \textit{We consider a WSI dataset $\mathcal{S}$ to consist of N whole slide images. Elements of $\mathcal{S}$ are tuples, i.e. $\mathcal{S}:=\{(s_1, y^{wsi}_1),..., (s_N, y^{wsi}_N)\}$, where $s_i$ and $y^{wsi}_i$ denotes the $i^{th}$ whole slide image and its disease class label.}

\noindent\textbf{Definition 2} (Patch Dataset). \textit{$i^{th}$ whole slide image $s_i$ consists of K tuples after preprocessing, i.e., $s_i$:=$\{(p_{ik}, y^{patch}_{ik})\}^{K}_{k=1}$. $p_{ik}$ and $y^{patch}_{ik}$ denote the extracted image patch and its label which is only known when the corresponding WSI is disease-negative, i.e., $y^{patch}_{ik}=0$ $\forall k$ when $y^{wsi}_i=0$. The number of image patches $K$ varies per slide $s_i$. We denote the encoded image patch $x_{ik}:=f(p_{ik})$ and $x_{i}:=\{x_{ik}\}^{K}_{k=1}$ where $f(\cdot)$ is the offline image encoder in Fig. \ref{fig1}}.

% to capture the common factors of variation within the normal patches by mapping all patches from normal bags into compact embeddings.

% \subsection{Revisiting Multiple Instance Learning for Whole Slide Image Classification}
\vspace{3mm}
\textbf{Revisiting Multiple Instance Learning for Whole Slide Image Classification.}
\label{sec:method1}
In general, as illustrated in Fig.\ref{fig1}, the WSI classification pipeline usually consists of two stages: 1) image encoder $f(\cdot)$, and 2) MIL module (Fig. \ref{fig1}-c). \cite{tu2022dual}. For the image encoder, most recent studies utilize offline pre-trained image encoder, e.g., using ImageNet, which can reduce the computational cost of model training and memory consumption \cite{tu2022dual, wang2022scl}. We use a self-supervised pre-trained ResNet-50 using ImageNet \cite{caron2020unsupervised} and DSMIL pre-trained ResNet-18 \cite{li2021dual} for Camelyon-16 and TCGA-Lung cancer classification tasks, respectively.

MIL module (Fig. \ref{fig1}-c) usually consists of two stages: 1) mapping function $g(\cdot)$ and instance embedding aggregator $h(\cdot)$, which, in recent popular attention-based MIL, aggregate each instance embedding using their respective attention score using querying function \cite{li2021dual, zhang2022dtfd, ilse2018attention}. In this study, we aim to constrain $g(\cdot)$ to yield compact embedding for all disease-negative patches.
 % and thus can learn more discriminative features for disease-negative and positive.

% \subsection{Compact Negative Embeddings with Debiased Bag Prior}
\textbf{Compact Negative Embeddings with Debiased Bag Prior.}
\label{sec:method2}
We hypothesize that both 1) learning compact negative instance representations from disease-negative bags and 2) reducing the `bag prior' \cite{ruff2018deep, ruff2019deep} can improve the predictive performance of the MIL algorithm. Note that Fig. \ref{fig1} depicts the exemplary `bag prior' caused by the staining protocol; homogeneous/heterogeneous staining protocol can lead to more similar/less similar image patch embedding, and it can be wrongly exploited by the MIL model and thus may create a spurious correlation between unnecessary image context and the label.

% To demonstrate our hypothesis, we first prepared our dataset using the offline image encoder $f(\cdot)$, which is addressed in Sec \ref{sec:method1}, to transform our patch dataset.

% \begin{table}[!ht]
%     \footnotesize
%     \centering
%     \renewcommand{\arraystretch}{0.92}
%     \caption{Data statistics with the number of subjects for mortality prediction, vasopressor need, and intubation need prediction tasks with modality missing information.}\label{tab1}
%     \begin{tabular}{l|ll|ll|ll}
%     \toprule
%         \multicolumn{7}{c}{\textbf{(a) Mortality Prediction}} \\  \hline
%         {} & \multicolumn{2}{c}{Training} & \multicolumn{2}{c}{Validation} & \multicolumn{2}{c}{Test} \\ \cline{2-7}
%         ~ & Positive & Negative & Positive & Negative & Positive & Negative \\ \hline
%         Patient Number & 3486 & 30870 & 430 & 3741 & 413 & 3873 \\ \hline
%         % \hspace*{3mm} Image: O, Text: O & 784 & 5338 & 106 & 597 & 101 & 652 \\ \hline
%         % \hspace*{3mm} Image: X, Text: O & 1072 & 9909 & 125 & 1210 & 131 & 1269 \\ \hline
%         % \hspace*{3mm} Image: O, Text: X & 80 & 1502 & 10 & 205 & 6 & 191 \\ \hline
%         % \hspace*{3mm} Image: X, Text: X & 1550 & 14121 & 189 & 1729 & 175 & 1761 \\ \hline
%         Image Missing Rate & 75.22\% & 77.84\% & 73.02\% & 78.56\% & 74.09\% & 78.23\% \\ \hline
%         Text Missing Rate & 46.76\% & 50.61\% & 46.28\% & 51.70\% & 43.83\% & 50.40\% \\\bottomrule
%     \end{tabular}

% \end{table}

% \documentclass{article}
% \usepackage{multirow}
\begin{table}[!t]
    \footnotesize
    \centering
    \renewcommand{\arraystretch}{1.15}
    \caption{Data statistics of Camelyon-16 and TCGA-Lung}\label{tab1}
    \begin{tabular}{lllll}
    \hline
        \multicolumn{2}{c}{\textbf{Dataset}} & \textbf{Training} & \textbf{Test} & \textbf{Total}  \\ \hline
        {\textbf{Camelyon-16}} & \# WSI & 270 & 129 & 399  \\ 
        ~ & \# Patch & 2,428,707 & 1,174,691 & 3,603,398  \\ \hline
        \textbf{TCGA-Lung} & \# WSI & 751 & 248 & 999  \\ 
        ~ & \# Patch & 12,048,148 & 3,156,655 & 15,204,803  \\ \hline
    \end{tabular}
\end{table}
\vspace{-3mm}

\begin{table}[t]
    \normalsize
    % \small
    \centering
    \tabcolsep=0.175cm
    \renewcommand{\arraystretch}{1.0}
    \caption{Performance comparison between baseline MIL models and MIL models with our proposed method.}\label{tab2}
    \begin{tabular}{ll|ll|ll}
    \hline  
        \multicolumn{2}{c}{\textbf{Dataset}} & \multicolumn{2}{c}{\textbf{Camelyon-16}} & \multicolumn{2}{c}{\textbf{TCGA-Lung}}  \\ \hline
        \multicolumn{2}{c}{\textbf{Method}} & AUC & ACC & AUC & ACC  \\ \hline
        ~ & ~ & 85.08 & 86.82 & 97.47 & 93.18  \\
        DSMIL\cite{li2021dual} & \multicolumn{1}{c|}{+Ours} & \textbf{88.11} & \textbf{87.75} & \textbf{97.56} & \textbf{93.84}  \\ 
        ~ & \multicolumn{1}{c|}{$\Delta$} & \textcolor{green!60!black}{{\footnotesize +3.03}} & \textcolor{green!60!black}{{\footnotesize +0.93}} & \textcolor{green!60!black}{{\footnotesize +0.09}} & \textcolor{green!60!black}{{\footnotesize +0.66}} \\ \hline
        
        ~ & ~ & 86.28 & 85.74 & 97.79 & 93.93  \\ 
        DTFD-MIL\cite{zhang2022dtfd} & \multicolumn{1}{c|}{+Ours} & \textbf{90.16} & \textbf{88.37} & \textbf{97.88} & \textbf{94.41}  \\ 
        $\scriptstyle{(AFS)}$ & \multicolumn{1}{c|}{$\Delta$} & \textcolor{green!60!black}{{\footnotesize +3.88}} & \textcolor{green!60!black}{{\footnotesize +2.63}} & \textcolor{green!60!black}{{\footnotesize +0.09}} & \textcolor{green!60!black}{{\footnotesize +0.48}} \\ \hline
        
        ~ & ~ & 82.61 & 86.2 & 96.4 & 89.19  \\ 
        ABMIL\cite{ilse2018attention} & \multicolumn{1}{c|}{+Ours} & \textbf{89.24} & \textbf{87.6} & \textbf{96.58} & \textbf{97.7}  \\
        $\scriptstyle{(Attention)}$ & \multicolumn{1}{c|}{$\Delta$} & \textcolor{green!60!black}{{\footnotesize +6.63}} & \textcolor{green!60!black}{{\footnotesize +1.4}} & \textcolor{green!60!black}{{\footnotesize +0.18}} & \textcolor{green!60!black}{{\footnotesize +8.51}} \\ \hline
        
        ~ & ~ & 85.71 & 86.36 & 96.53 & \textbf{97.38}  \\ 
        ABMIL\cite{ilse2018attention} & \multicolumn{1}{c|}{+Ours} & \textbf{88.77} & \textbf{87.6} & \textbf{96.99} & 90.24  \\
        $\scriptstyle{(GatedAttention)}$ & \multicolumn{1}{c|}{$\Delta$} & \textcolor{green!60!black}{{\footnotesize +3.06}} & \textcolor{green!60!black}{{\footnotesize +1.24}} & \textcolor{green!60!black}{{\footnotesize +0.46}} & \textcolor{red!90!black}{{\footnotesize -7.14}} \\ \hline
    \end{tabular}
\end{table}

Considering that all patches from the negative slide are disease-negative and from the positive slide are mostly negative as well \cite{li2021dual}, we were motivated by the approach from the anomaly detection community to impose a compact description of normal data \cite{ruff2018deep, ruff2019deep}. To this end, we design a supplementary loss function to minimize the standard deviation of patch embeddings $g(x_i)$ when $y^{wsi}_i=0$ (negative WSIs) as well as to synchronize the center of patch embeddings from every negative slide. Specifically, we used a single linear projection layer $l(\cdot)$ to transform $g(x_i)$ into a $M$-dimension representation and impose compact embeddings for all negative slides (we used $M=128$ throughout the study). We also jointly train $M$-dimension distribution center of the disease-negative instance embedding $\hat{\mu}$ to reduce the `bag prior' by minimizing the standard deviation of instance embeddings from all negative WSI using a single distribution center. The equation can be shown in Eq. \ref{eq:1}:

% feeding a compact negative embedding for MIL aggregator $h(\cdot)$

\begin{equation}\label{eq:1}
\mathcal{L}_{neg} = \frac{1}{M}\sum^M_m{\sqrt{\frac{1}{K-1}\sum_{k=1}^{K}{(l(g(x_{ik}))-\hat{\mu})^{2}}}}
\end{equation}

Note that $\hat{\mu}$ is a learnable parameter shared by all WSI slides to synchronize the center of negative embedding. Our algorithm is not bound to a binary classification task but is applicable to multi-class classification tasks as well. In a multi-class classification scenario, we learn multiple embedding centers with multiple linear projection layers $l(\cdot)$ (equal to the class number). To avoid degenerate solutions, e.g., zeros for all parameters in $l(\cdot)$, we increase the standard deviation when the standard deviation is smaller than the threshold $thr$. The equation can be shown in Eq. \ref{eq:2}:

\begin{equation}\label{eq:2}
\begin{aligned}
    \mathcal{L}_{pos} = &\frac{1}{M}\sum^M_m ReLU \bigg( \\
    &thr - \sqrt{\frac{1}{K-1}\sum_{k=1}^{K}{(l(g(x_{ik}))-\hat{\mu})^{2}}} \bigg)
\end{aligned}
\end{equation}

As a result, the overall loss can be described as below. Throughout this paper, we call our method CDNE, i.e. compact and debiased negative embedding:

\begin{equation}\label{eq:3}
\mathcal{L}_{overall} = \mathcal{L}_{MIL} + \lambda_{neg}\mathcal{L}_{neg} + \lambda_{pos}\mathcal{L}_{pos}
\end{equation}

\begin{figure}[t]
\centering
\includegraphics[width=0.48\textwidth]{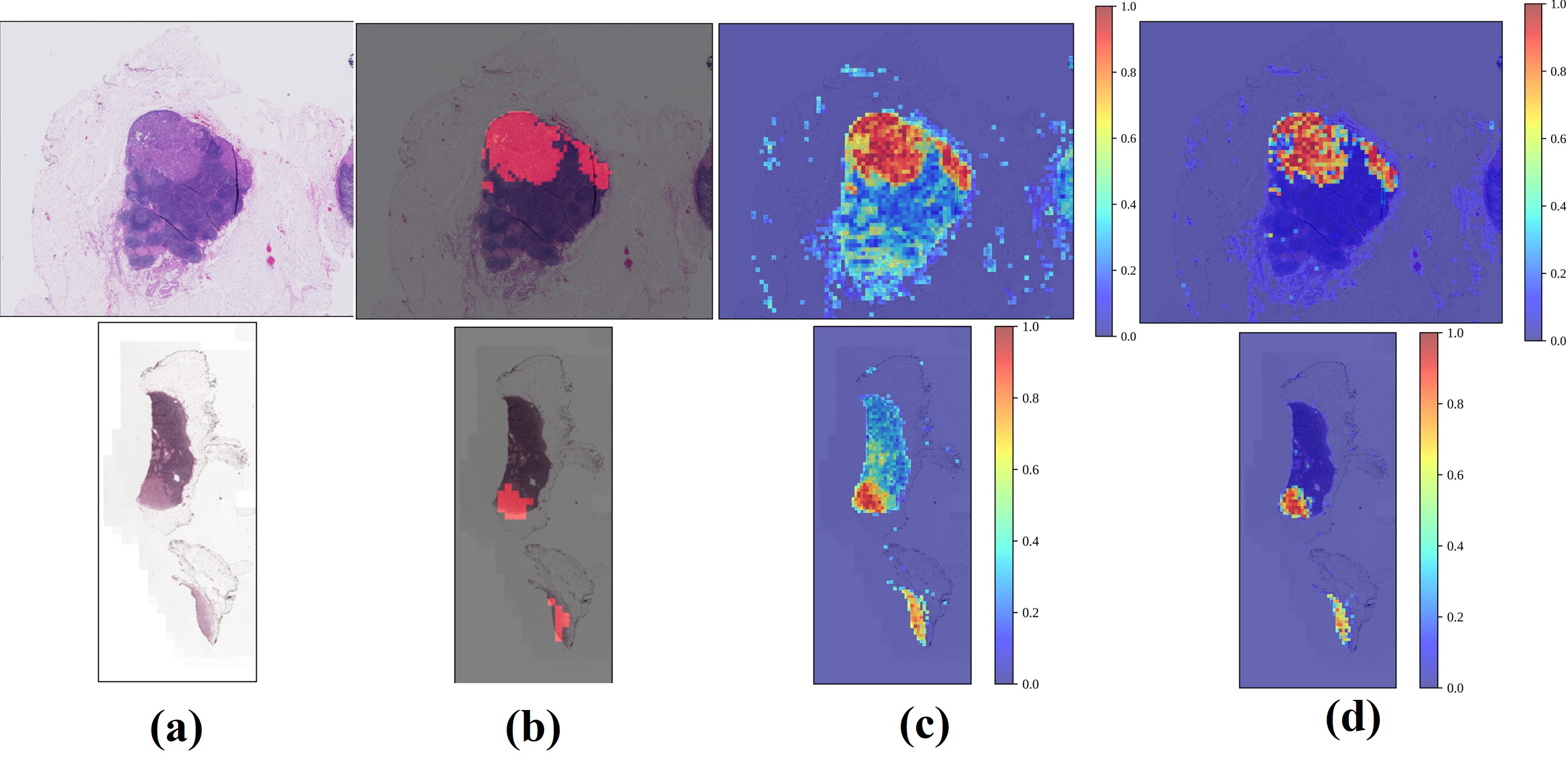}
\caption{Attention map comparison with/without our method on ABMIL (Gated Attention) with 2 test WSIs. (a) WSI without annotation; (b) WSI with annotation; (c) heatmap from ABMIL (Gated Attention) baseline; (d) heatmap from Ours + ABMIL (Gated Attention) baseline}
\label{fig3}
\end{figure}

\section{Experiment}
% \subsection{Dataset}
\textbf{Dataset.}
% \vspace{-2mm}
In this study, we employ two datasets for histopathological analysis: Camelyon-16 and TCGA-Lung (Table \ref{tab1}). The Camelyon-16 dataset comprises 399 WSIs, with 270 allocated for training and 129 for testing, totaling over 3.6 million patches. On the other hand, the TCGA-Lung dataset features 999 WSIs, with 751 designated for training and 248 for testing, culminating in more than 15.2 million patches. As a preprocessing, we segmented both datasets into smaller patches with a fixed mpp of 0.5 and disregarded background patches. For all performance evaluations, we use the averaged area under the receiver operating characteristic (AUROC) and prediction accuracy from 5-fold cross-validation (CV) using a test set.

% \subsection{CDNE increases the performance of MIL}
\textbf{CDNE increases the performance of MIL.}
% \vspace{-2mm}
We implemented four recent MIL algorithms and examined the efficacy of our approach on top of them: DSMIL\cite{li2021dual}, DTFD-MIL\cite{zhang2022dtfd}, ABMIL-Attention, and ABMIL-Gated Attention \cite{ilse2018attention}. To assess the performance of our method, we benchmark the performance of these MIL models both with and without employing our CDNE. As illustrated in Table \ref{tab2}, the application of CDNE significantly boosts both the AUROC and accuracy metrics for the Camelyon-16 dataset. In the case of the TCGA-Lung dataset, where baseline MIL models already demonstrate strong performance, CDNE still manages incremental improvements in predictive accuracy, with the exception of the ABMIL-Gated Attention in accuracy metric. Note that both pulling through negative embeddings (eq.\ref{eq:1}) and pushing through positive embeddings (eq.\ref{eq:2}) contribute to the performance increase, and there exists an optimal loss weight for both losses as described in Table. \ref{tab4}.

\begin{table}[!ht]
    \small
    % \normalsize
    % \large
    \centering
    \renewcommand{\arraystretch}{1.0}
    \caption{Performance comparison between baseline DSMIL\cite{li2021dual} and DSMIL with various SSL methods.}\label{tab3}
    \begin{tabular}{l|llll}
    \hline  
        ~ & \multicolumn{4}{c}{\textbf{Dataset}} \\ \cline{2-5}
        \multicolumn{1}{c|}{\textbf{Method}} & \multicolumn{2}{c}{\textbf{Camelyon-16}} & \multicolumn{2}{c}{\textbf{TCGA-Lung}}  \\ \cline{2-5}
        ~ & AUC & ACC & AUC & ACC  \\ \hline
        \multicolumn{1}{c|}{DSMIL\cite{li2021dual}} & 85.08 & 86.82 & 97.47 & 93.17 \\ \hline
        \multicolumn{1}{c|}{+Ours} & \textbf{88.11} & \textbf{87.75} & \textbf{97.56} & \textbf{93.84}\\
        \multicolumn{1}{c|}{$\Delta$} & \textcolor{green!60!black}{{\footnotesize +3.03}} & \textcolor{green!60!black}{{\footnotesize +0.93}} & \textcolor{green!60!black}{{\footnotesize +0.09}} & \textcolor{green!60!black}{{\footnotesize +0.67}}  \\ \hline
        \multicolumn{1}{c|}{+DivDis\cite{lee2022diversify}} & 82.29 & 85.11 & 97.5 & 92.98\\ 
        \multicolumn{1}{c|}{$\Delta$} & \textcolor{red!90!black}{{\footnotesize -2.79}} & \textcolor{red!90!black}{{\footnotesize -1.71}} & \textcolor{green!60!black}{{\footnotesize +0.03}} & \textcolor{red!90!black}{{\footnotesize -0.19}}  \\ \hline
        \multicolumn{1}{c|}{+Psuedo-Label\cite{myronenko2021accounting}} & 77.36 &  81.6 & 97.16 & 92.1\\ 
        \multicolumn{1}{c|}{$\Delta$} & \textcolor{red!90!black}{{\footnotesize -7.72}} & \textcolor{red!90!black}{{\footnotesize -5.22}} & \textcolor{red!90!black}{{\footnotesize -0.31}} & \textcolor{red!90!black}{{\footnotesize -1.07}}  \\ \hline
    \end{tabular}
\end{table}

% \begin{table}[!ht]
%     \footnotesize
%     % \normalsize
%     % \large
%     \centering
%     \renewcommand{\arraystretch}{1.0}
%     \caption{Performance comparison between baseline DSMIL and DSMIL with semi-supervised learning methods. The table shows results in Test AUROC.}\label{tab3}
%     \begin{tabular}{lll}
%     \hline  
%         {} & \textbf{Camelyon-16} & \textbf{TCGA-Lung}  \\ \hline
%         DSMIL & 85.08 & 97.47  \\ \hline
%         DSMIL + ours & \textbf{88.11}\textcolor{green!60!black}{{\footnotesize +3.03}} & \textbf{97.56}\textcolor{green!60!black}{{\footnotesize +0.09}}\\ \hline
%         DSMIL + DivDis & 82.29\textcolor{red!90!black}{{\footnotesize -2.79}} & 97.5\textcolor{green!60!black}{{\footnotesize +0.03}}\\ \hline
%         DSMIL + Psuedo-Label\ref{myronenko2021accounting} & 77.36\textcolor{red!90!black}{{\footnotesize -7.72}} & 97.16\textcolor{red!90!black}{{\footnotesize -0.31}}\\ \hline
%     \end{tabular}
% \end{table}

\begin{table}[!ht]
    \small
    % \normalsize
    \centering
    \caption{Ablation study on $\lambda_{Negative}$ and $\lambda_{Positive}$ with DSMIL+ours in Camelyon-16 test set.}\label{tab4}
    \begin{tabular}{l|lllll}
    \hline
        \textbf{$\lambda_{Negative}$} & \textbf{0}  & \textbf{1}  & \textbf{10}  & \textbf{100}  & \textbf{1000}  \\ \hline
        AUC  & 85.87  & 86.80  & 88.11  & 70.71  & 64.95  \\ \hline
        ACC  & 87.44  & 86.98  & 87.75  & 69.77  & 61.71  \\ \hline
        \textbf{$\lambda_{Positive}$} & \textbf{0}  & \textbf{0.3}  & \textbf{3}  & \textbf{30}  & \textbf{300}  \\ \hline
        AUC  & 86.93  & 81.78  & 88.11  & 85.0  & 84.45  \\ \hline
        ACC & 87.13 & 83.26 & 87.75 & 86.67 & 86.51 \\ \hline
    \end{tabular}
\end{table}

% \subsection{CDNE outperforms other semi-supervised method}
\textbf{CDNE outperforms other semi-supervised method.}
Our approach can be categorized as semi-supervised learning (SSL) in that it utilizes the implicit patch labels of disease-negative WSI; all patches from the negative slides are negative. Consequently, we compare the performance gain with other semi-supervised approaches. Specifically, we compare the predictive performance of MIL when attached with CDNE (ours), Myronenko \textit{et al.} \cite{myronenko2021accounting}, and DivDis \cite{lee2022diversify} in Dsmil setting \cite{li2021dual}. As described in Table 3, our approach outperforms other alternatives that can utilize the patch label (negative) from negative WSI.

% \subsection{CDNE de-biases negative embeddings and learn better discriminative feature to distinguish positive and negative WSI.}
\textbf{CDNE de-biases negative embeddings and learn better discriminative feature to distinguish positive and negative WSI.}
% \vspace{-2mm}
As shown in Fig. \ref{fig2}-(a), our method effectively reduces the inter-slide bias when attached to the existing MIL algorithm. Moreover, our compact negative embedding approach creates distinct standard deviations between positive and negative WSIs as described in figure \ref{fig2}-(b,c). Our compact negative embedding method also leads to a more accurate attention map as described in Fig. \ref{fig3}. Note that our method significantly reduces the false positives in Fig. \ref{fig3}.

% \subsection{Compact and debiased negative embedding creates more precise attention map.}
% % \vspace{-2mm}
% As shown in Fig. \ref{fig2}-(a), our method effectively reduces the inter-slide bias when attached to the existing MIL algorithm. Moreover, our compact negative embedding approach creates distinct standard deviations between positive and negative WSIs as described in figure \ref{fig2}-(b,c).

\section{Conclusion}
We hypothesize and demonstrate that reducing the inter-slide bias and learning a compact negative embedding improve the predictive performance of the MIL algorithm. To this end, we use the implicit labels of negative WSIs and compress their instance embeddings to a single negative center. Through experiments, we confirm that our method effectively reduces the inter-slide bias and creates compact negative embeddings by showing the standard deviation difference between positive and negative WSIs (Fig. \ref{fig2}).

\vfill\pagebreak

% \section{REFERENCES}
% \label{sec:refs}

% List and number all bibliographical references at the end of the
% paper. The references can be numbered in alphabetic order or in
% order of appearance in the document. When referring to them in
% the text, type the corresponding reference number in square
% brackets as shown at the end of this sentence \cite{C2}. An
% additional final page (the fifth page, in most cases) is
% allowed, but must contain only references to the prior
% literature.

\bibliographystyle{IEEEbib}
\bibliography{strings,refs}

\end{document}